\documentclass[runningheads]{llncs}
\usepackage{amsfonts}
\usepackage{makecell}
\usepackage{adjustbox}
\usepackage{graphicx}
\usepackage{algorithm}
\usepackage{algorithmic}
\usepackage{makecell}
\usepackage{colortbl}
\definecolor{Gray}{gray}{0.9}
\usepackage{multirow}
\usepackage{booktabs}
\begin{document}

\title{RoomDiffusion: A Specialized Diffusion Model in the Interior Design Industry}
\author{Zhaowei Wang, Ying Hao, Hao Wei, Qing Xiao, Lulu Chen, Yulong Li, \\Yue Yang\thanks{Corresponding author: yangyue092@ke.com}, Tianyi Li}
\institute{Beike}

\maketitle              
\begin{abstract}
\vspace{-10mm}
\begin{sloppypar}

Recent advancements in text-to-image diffusion models have significantly transformed visual content generation, yet their application in specialized fields such as interior design remains underexplored. 
In this paper, we present \textbf{RoomDiffusion}, a pioneering diffusion model meticulously tailored for the interior design industry. 
To begin with, we build from scratch a whole data pipeline to update and evaluate data for iterative model optimization. 
Subsequently, techniques such as multi-aspect training, multi-stage fine-tune and model fusion are applied to enhance both the visual appeal and precision of the generated results.
Lastly, leveraging the latent consistency Distillation method, we distill and expedite the model for optimal efficiency.
Unlike existing models optimized for general scenarios, RoomDiffusion addresses specific challenges in interior design, such as lack of fashion, high furniture duplication rate, and inaccurate style.
Through our holistic human evaluation protocol with more than 20 professional human evaluators, RoomDiffusion demonstrates industry-leading performance in terms of aesthetics, accuracy, and efficiency, surpassing all existing open source models such as stable diffusion and SDXL. 
\end{sloppypar}
\end{abstract}
\section{Introduction}
For text-to-image diffusion models~\cite{ho2020denoising,dhariwal2021diffusion,song2020denoising}, the realm of interior design stands as a splendid arena for application. 
In traditional interior design processes, people often find themselves in prolonged and costly exchanges with professional designers; another potential challenge is that many individuals often lack clarity about their needs, leading to a disorganized and inefficient design process.
With the growing popularity of diffusion models, these issues appear to be effectively resolved. 
People can leverage text-to-image diffusion models to quickly explore a vast array of design ideas, thereby gaining the inspiration they seek. 
Additionally, such models can assist professional designers in rapidly generating designs, thus enhancing their efficiency in the workflow.

\begin{figure}
\centering
\includegraphics[width=0.90\textwidth]{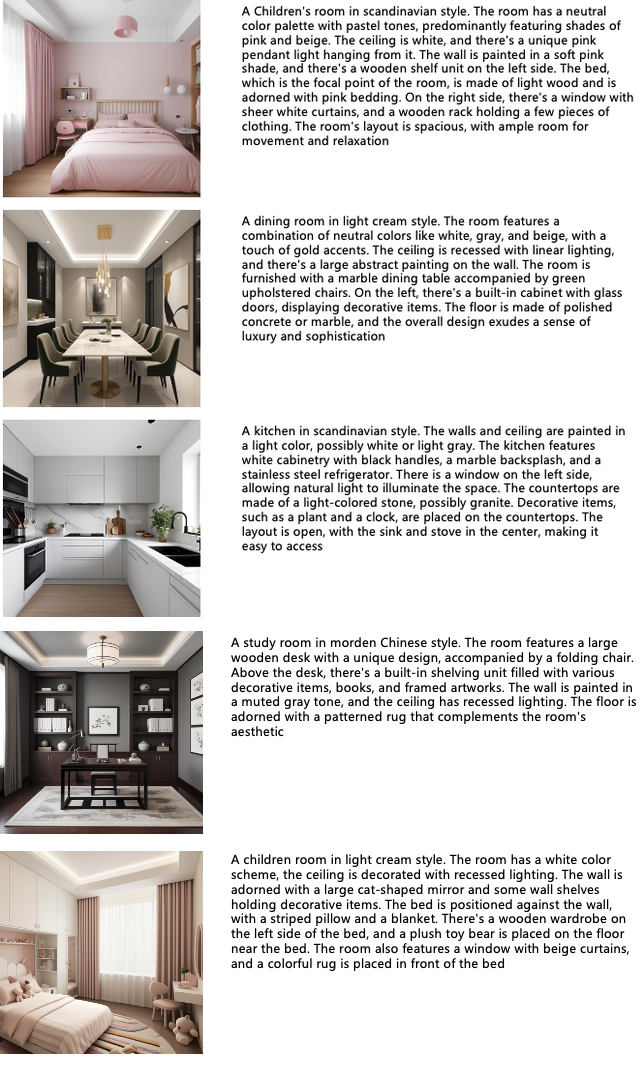}
\caption{RoomDiffusion can generate images following long text prompts.}
\label{fig3}
\end{figure}

\begin{figure}
\centering
\includegraphics[width=\textwidth]{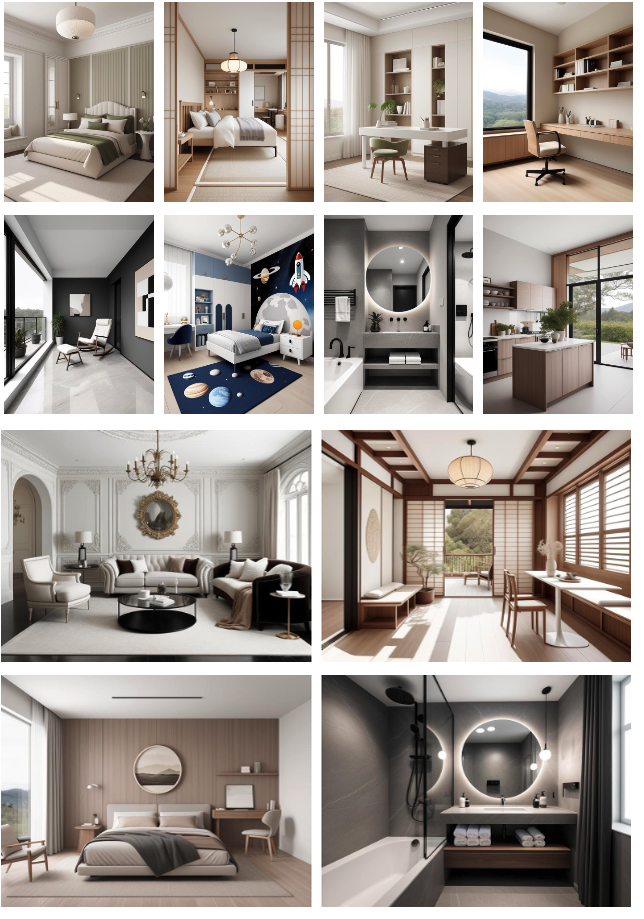}
\caption{RoomDiffusion can generate images in various resolutions.}
\label{fig4}
\end{figure}

\begin{figure}
\centering
\includegraphics[width=\textwidth]{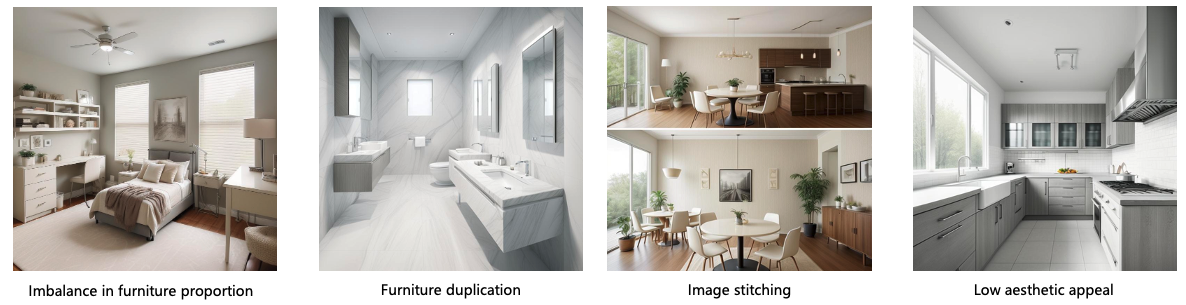}
\caption{The shortcomings of open source models in interior design}
\label{fig1}
\end{figure}

However, existing open-source text-to-image diffusion models like stable diffusion~\cite{rombach2022high} and SDXL~\cite{podell2023sdxl} are primarily designed to cater to general applications, and their performance in specialized fields is somewhat lacking. 
Due to the low signal-to-noise ratio in training data and the limited quantity of indoor scene data, open-source models often exhibit issues such as repetitive furniture, outdated styles, imbalanced furniture proportions, and disjointed compositions, as illustrated in Figure.~\ref{fig1}.

In this report, we introduce RoomDiffusion to address the aforementioned issues and present the entire process of building RoomDiffusion: (1) we created a massive dataset comprising tens of millions of indoor scene images sourced from various channels. Building upon this foundation, we established a comprehensive system for evaluating the quality of indoor images, assigning 19 labels to each image. Some of these labels were utilized to filter out low-quality images, while others formed the textual components of the dataset. (2) we divide the filtered images into several buckets based on their resolution, and randomly extract data from different buckets during model training. At the same time, we using image resolution as a conditional input to control the training process. (3) we selected high-quality images to create a high-precision dataset, then further fine-tuned the model obtained in the previous step based on this dataset, as done in~\cite{dai2023emu}. (4) we selected several outstanding open-source text-to-image diffusion models for model fusion, such as EpicRealism~\cite{epic} and Realistic Vision~\cite{rv51}. (5)  we applied latent consistency Distillation(LCD)~\cite{luo2023latent} to enhance the model's inference speed.

To comprehensively evaluate the performance of RoomDiffusion, we combined automated metrics with human assessment. 
For automated metrics, we assessed aesthetic quality, CLIP score~\cite{clip}, Frechet Inception Distance(FID)~\cite{fid}, and several other indicators. 
As for human evaluation, we collaborated with over 20 professional evaluators to establish a rational evaluation framework, focusing on aesthetic appeal, image-text alignment, and spatial coherence through good-same-bad(GSB) assessment.
RoomDiffusion demonstrated a leading advantage across all automated metrics. 
In human evaluations, RoomDiffusion surpassed the best open-source models with an 70\% win rate across multiple dimensions, demonstrating its superior performance. 

\begin{figure}
\centering
\includegraphics[width=\textwidth]{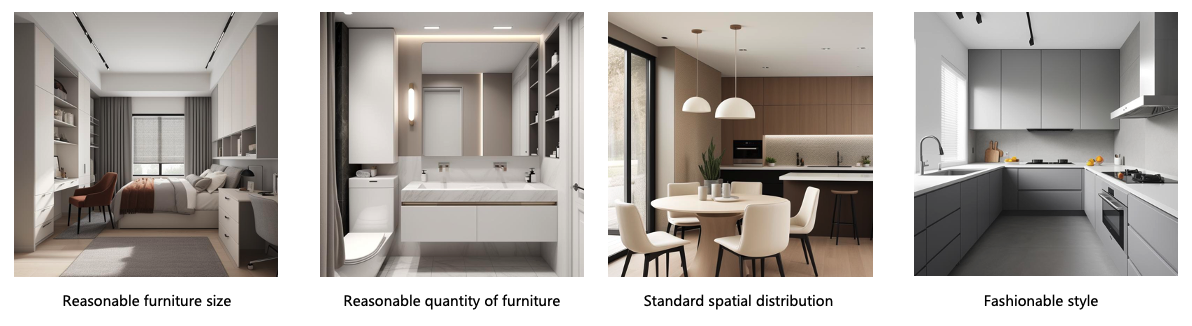}
\caption{The generation results of RoomDiffusion in interior design}
\label{fig2}
\end{figure}

\section{Method}
\begin{figure}
\centering
\includegraphics[width=0.8\textwidth, keepaspectratio]{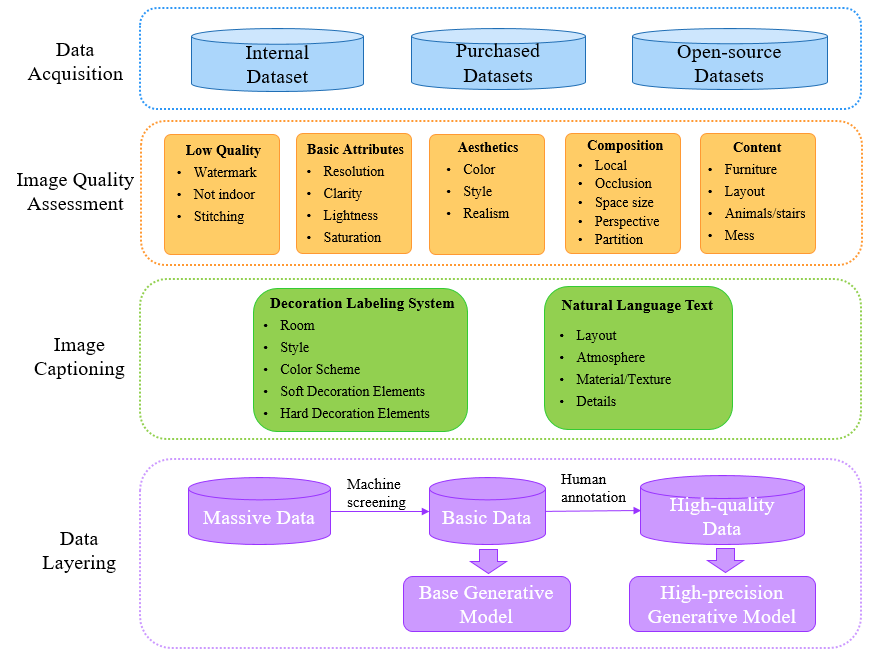}
\caption{The architecture diagram of the data pipeline.}
\label{fig2-1}
\end{figure}

In this chapter, we will provide a comprehensive overview of the entire construction process of RoomDiffusion.  
In Section 2.1, we will introduce our data pipeline, with its overall structure illustrated in Figure.~\ref{fig2-1}. Then, in Section 2.2, we will describe all the technologies used in RoomDiffusion, such as Multi-aspect training and LCD, with the complete workflow depicted Figure.~\ref{fig2-3}.

\subsection{Data Pipeline}

\subsubsection{Raw Data Acquisition}
In order to build a leading-edge and high-performance text-to-image diffusion model, a comprehensive and well-curated dataset is indispensable.
Leveraging years of experience in the residential sector, we have amassed a substantial collection of high-quality interior design renderings.
Additionally, we have augmented the diversity of our training data through external data procurement and open-source downloads. These efforts have culminated in the creation of a dataset comprising tens of millions of decoration renderings.
\subsubsection{Image Quality Assessment} 
After obtaining the raw data, we have identified common issues in the images and developed a set of image quality labeling system to evaluate the quality of the images. Currently, this system consists of 5 primary labels and 19 secondary labels. The specific primary labels are explained below:
\begin{enumerate}
\item \textbf{Low Quality}: Assessing whether the image is usable, including criteria such as whether the image is not an indoor rendering, stitched, or watermarked.
\item \textbf{Basic Attributes}: Including image resolution, clarity, brightness and saturation.
\item \textbf{Aesthetics}: 
Evaluating the aesthetic quality of the image, including assessing whether there are issues such as color mismatches, outdated styles, or a lack of realism.
\item \textbf{Composition}: Assessing whether the proportions of the main objects in the image are reasonable and identifying issues such as excessive focus on a specific area, occlusion, or incorrect shooting angles.
\item \textbf{Content}: Evaluating whether the content of the image is reasonable, such as whether the number of key furniture items is appropriate, or whether it includes people or animals.
\end{enumerate}

\begin{figure}
\centering
\includegraphics[width=0.7\textwidth, keepaspectratio]{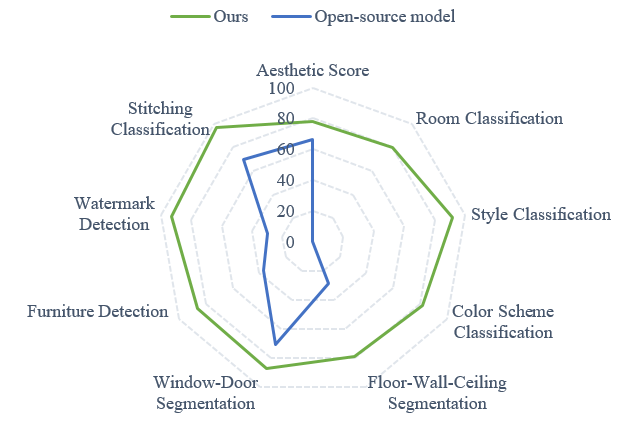}
\caption{Comparison of performance between our domain model and open-source models~\cite{ocr,beauty,sam,liu2023grounding}.}
\label{fig2-2}
\end{figure}

\begin{figure}
\centering
\includegraphics[width=\textwidth]{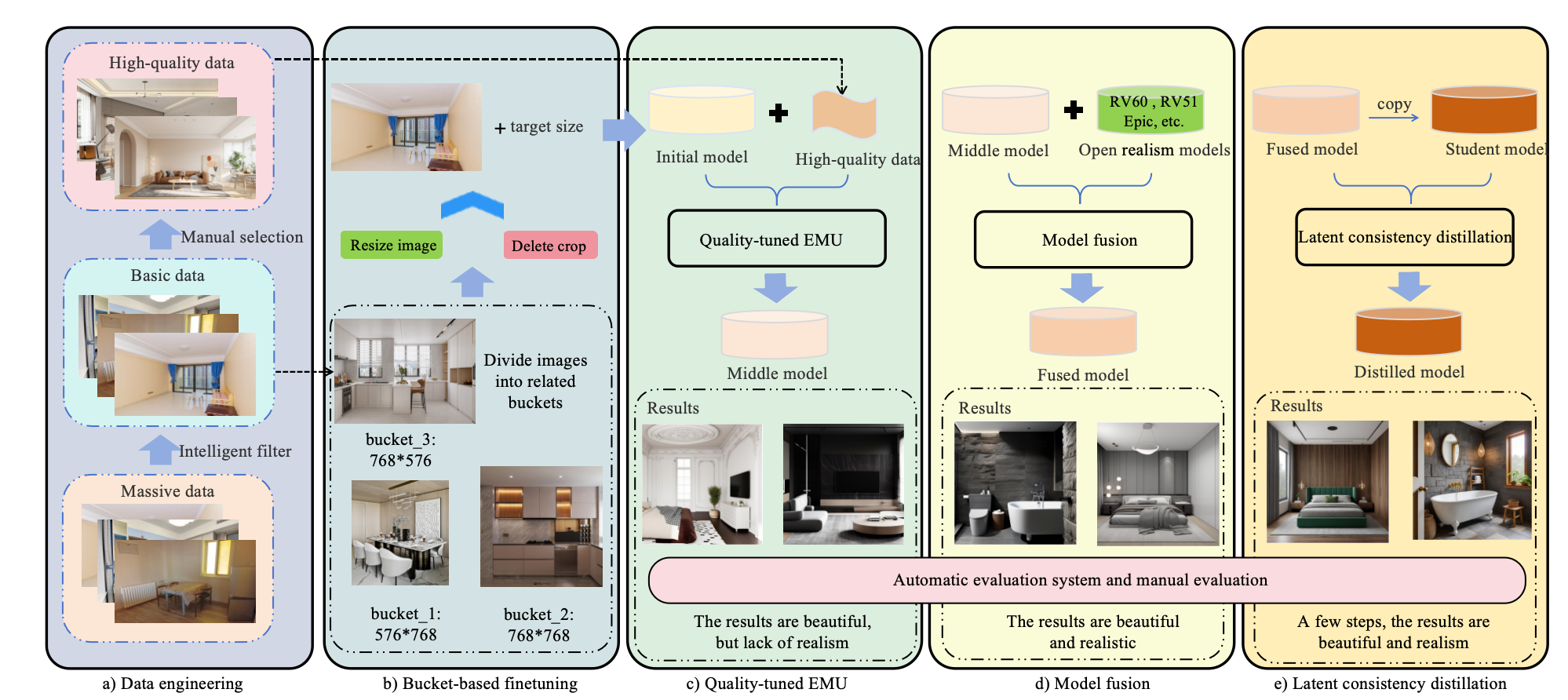}
\caption{The complete workflow of RoomDiffusion }
\label{fig2-3}
\end{figure}

To obtain accurate image labels as described above, we have developed over ten domain models, including watermark detection,  stitching classification, aesthetic scoring, indoor segmentation, and indoor detection. These models have demonstrated superior performance compared to existing open-source models~\cite{ocr,beauty,sam,liu2023grounding}, as shown in Figure.~\ref{fig2-2}.
By applying the models, we can set corresponding discrimination rules for each label. Therefore, we can utilize these models to perform preliminary screening on large-scale datasets, eliminating all low-quality data, and selecting high-quality images for model training.

\subsubsection{Image Captioning} 
We have constructed two forms of image descriptions, one based on label systems and the other based on natural language text.
\begin{enumerate}
\item \textbf{Labeling system: }
We have constructed a labeling system for home decoration scenes, consisting of 5 primary labels, 98 secondary labels, and over three hundred tertiary labels. The primary labels encompass five aspects: room, style, color scheme, soft decoration elements, and hard decoration elements. In order to provide comprehensive labels for each image, we trained a classification model to categorize the room, style, and color scheme of the images. Additionally, we employed a detection model to detect the presence of soft and hard decoration elements within the images. As a result, we are able to accurately interpret the elements depicted in the images.

\item \textbf{Natural language text}
\newline Since label-based image descriptions lack details such as furniture material, color, and spatial relationships, we further enhance the image descriptions using natural language text. To ensure the accuracy and richness of the image descriptions, we conducted a detailed comparison of the image caption capabilities of different language models. Ultimately, we chose to utilize CogVLM-chat~\cite{wang2023cogvlm} and GPT-4V(ision)~\cite{achiam2023gpt} for batch production of text descriptions for images. We designed prompts to guide the models in describing various aspects of the images, including room, style, walls, ceilings, floors, decoration status, furniture, and layout, in order to obtain comprehensive descriptions of the images.

\end{enumerate}

In conclusion, we combine the image quality labels, home decoration labels, and natural language text to obtain the most accurate and detailed descriptions of the images. The format of the textual descriptions is as follows: "[room] + [style] + [quality labels (watermark, clarity, etc.)] + [furniture] + [natural language text]."
Notably, CLIP can typically only accept up to 77 tokens, making it unable to handle most image captions. We split long captions into multiple short captions, encode them separately with CLIP, and then concatenate the results before feeding them into the UNet. This approach enables RoomDiffusion to understand long texts.

\subsubsection{Data Layering}
To fully exploit the value of large-scale datasets, we layer the data based on different qualities and quantities and apply them at different stages of model training.

For instance, we employ the aforementioned image quality indicators to conduct preliminary screening of the images, and combining them with text generated by the CogVLM-chat model to create a dataset comprising millions of image-text pairs. This dataset is utilized for training the generation model, aiming to enhance the model's ability to generate outputs with improved aesthetics, semantic control, and coherence. Our objective is to maximize the expansion of the model's generative boundaries and diversity. Furthermore, we utilize a dataset of hundreds of thousands of images manually selected through screening, combined with text generated by the GPT-4V(ision) model. This combined dataset is employed to fine-tune the generated distribution, thereby narrowing down the model's generated distribution to a sample space of higher quality.

\subsection{Model Training}
We have designed a new training pipeline to enhance the model's generation performance, improving its aesthetic quality, rationality, and semantic control capability. This pipeline consists of four main components: Multi-aspect training, Multi-stage fine-tune, Model Fusion, and LCD, the entire workflow is shown in the figure.~\ref{fig2-2}.

\subsubsection{Multi-aspect training}
In the context of indoor scene, the resolution plays a crucial role in the results generated by models. Currently, most open-source models generate images at a single resolution, which falls far short of meeting the practical demand for larger and a variety of image sizes. Additionally, in the domain of interior decoration characterized by complexity and a plethora of elements, various unreasonable issues are prone to arise, significantly affecting the reference value of generated images in practical applications.
For instance, if we train the model at a specific resolution, it performs relatively well when inferring at or near that resolution. However, when the inference resolution is larger than the trained resolution, it is prone to issues such as distorted or missing furniture. Conversely, when the inference resolution is smaller, the generated image may be blurry and may exhibit insufficient or distorted furniture details.

To ensure the model performs exceptionally well across various resolutions, we employed a multi-aspect training approach. This method not only enhances the model's performance at different resolutions but also increases its stability and flexibility in generative tasks.
We partition the data into multiple buckets based on different aspect ratios, assigning each training image to the bucket with the closest aspect ratio. In each iteration, data is sampled from a randomly selected bucket for training. Additionally, the model receives the resolution of the target bucket as a condition to control image generation.
To ensure that the semantic alignment between images and text is maximized, we only perform resizing in the image preprocessing step during training and eliminate all cropping operations. Since the images are placed in buckets with similar aspect ratios, they are not excessively compressed or stretched, and the proportions of objects remain relatively normal.

\subsubsection{Multi-stage fine-tune}
After training on a large interior design dataset, our model has developed strong generative capabilities for indoor scenes. However, its aesthetic quality still requires improvement. To address this, we fine-tuned our model on a smaller, but higher-quality image dataset, as done in \cite{dai2023emu}.

To obtain exceptionally high-quality data, we built a data cleaning pipeline: 
(i) the raw data is initially filtered by the powerful image quality system, aiming to automatically filter out high-quality images from massive data and greatly reduce labor costs. Data volume reduced from tens of millions to 100k. 
(ii) We will divide the image beauty into 5 levels from multiple perspectives such as decoration colors, furniture selection, hard decoration design, etc., ranging from 1 to 5. The larger the number, the higher the standard beauty. A number of designers with interior design experience are trained to rate the beauty of images. Each picture will pass through multiple professional designers to finally obtain the average beauty value, and the top 10\% of the pictures with the highest score will be retained. 
(iii) To obtain corresponding high-quality prompts, we use  GPT-4V(ision) to produce detailed prompts for top 10\% images, and then manually checked the detailed prompts. 
Finally, we obtained 5,000 exceptionally high-quality image-text pairs. After training the model as described in the previous section, we continued fine-tuning it on these image-text pairs for 10,000 steps with a learning rate of 1e-6.
Through experimental observation, this method significantly enhances the aesthetic quality and texture of the images generated by the model.

\subsubsection{Model Fusion}
Although our model has shown significant improvements in aesthetics, semantic control, and coherence, it still has some issues. Due to the high proportion of rendered images in our training data, the results generated by our model lack realism.
To address this issue, we employed model fusion techniques widely used in the open-source community.
By integrating our model with those renowned for realism in the community, we can enhance the authenticity and detail of the generated images. 


However, since most open-source models are primarily designed for single-resolution image inference, the fusion process, while enhancing aesthetics and realism, still introduces issues such as duplicated furniture and image stitching artifacts. To address this, we initially conducted bucket fine-tuning on the open-source models using a small dataset and a low learning rate. This fine-tuning process ensured that the models maintained their realism while adapting to the generation of multi-scale images. Finally, we fused the fine-tuned models to achieve a balance between realism and coherence.

\subsubsection{LCD}
Latent Consistency Distillation (LCD) aims to efficiently distill the pre-trained classifier-free~\cite{ho2022classifier} guided diffusion models. LCD directly predicts the solution of such ODE in latent space, requiring only a few iterations, resulting in rapid, high-fidelity sampling. We attempt to apply LCD to accelerate our RoomDiffusion and observe some inspiring properties: (i). The teacher model and data quality jointly affect the performance of student models. (ii). The LoRA-based~\cite{hu2021lora} distillation scheme performs worse than the model-based scheme. (iii). CFG scale and batch size are the two critical aspects of the LCD distilling process. (iv). The distilled model may lose few high-frequency components during the sampling process. 

To alleviate these properties of the LCD method, we select the best-performing teacher model and the high quality data. We carefully tune the CFG scale, batch size and optimizer in the training process to encourage the student model to accurately imitate the teacher model. Furthermore, in order to solve the problem of missing high-frequency components, we use high-quality data to distill the open source photorealistic stable diffusion models and perform model fusion with our student model in proportion.
Finally, we successfully reduced the model inference time to one-third of the original while maintaining consistent performance across all metrics.

\section{Evaluation Protocol}
Traditional evaluation criteria for text-to-image generation generally include aspects such as text-image consistency, aesthetic quality of images, and content coherence. However, these metrics are insufficient for providing a comprehensive assessment of models in decoration scenario. 
On the one hand, they fail to identify key issues such as repetitive furniture, mixed styles, and poor fidelity of generated images; on the other hand, certain commonly used metrics may exhibit distortion. 
For instance, some existing aesthetic models tend to assign higher scores to images with abundant furniture and intricate patterns, while assigning lower scores to images with minimalist interior design styles. 
Hence, it is essential to revamp the evaluation process and develop additional metrics to ensure the credibility of the results.

\subsection{Evaluation Process}
We have designed a dual evaluation mechanism comprising two steps.The first step is to use automated evaluation metrics to quickly assess the model's performance during the iterative process. If over 70\% of the metrics show improvement, the evaluation moves to the next step, which involves human evaluation using the GSB method. This approach not only conserves human resources but also ensures more reliable results.

\subsection{Automated Evaluation Metrics}
We measure the performance of our model from multiple dimensions, which can be mainly divided into visual appeal and image-text consistency. Visual appeal includes Fréchet inception distance (FID), and aesthetic score (AS). Image-text consistency include CLIP score (CS) and fine-grained metrics, such as soft-decoration follow rate (SFR), style accuracy (SA), hard-decoration follow rate (HFR) and furniture repetition rate(FRR).
\begin{itemize}
    \item[•]\textbf{Fréchet inception distance}: FID is used to assess the quality of images created by a generative model. It has been used to measure the quality of many SOTA generative models.
    \item[•]\textbf{Aesthetic score}: The common aesthetic score model is inconsistent with human subjective judgment in the decoration scenario. We train an aesthetic scoring model based on images annotated by professional designers to evaluate the image aesthetic scores.
    \item[•]\textbf{CLIP Score}: is a reference free metric that can be used to evaluate the correlation between a generated caption for an image and the actual content of the image.
    \item[•]\textbf{Soft-decoration follow rate}: Soft-decoration follow rate statistics the generation accuracy of 49 important furniture.
    \item[•]\textbf{Style accuracy}: SA statistics the generation accuracy of 8 popular styles.
    \item[•]\textbf{Hard-decoration follow rate}: Hard-decoration follow rate statistics the generation accuracy of 10 common floors and ceilings. 
    \item[•]\textbf{Furniture repetition rate}: In the results generated by diffusion model, there are often unreasonable repetitions of furniture, such as two toilets in a bathroom or two double beds in a bedroom. Therefore, it is necessary to quantify this phenomenon.
\end{itemize}

\subsection{Manual evaluation}
To further enhance the confidence in the evaluation process, we conducted a manual assessment. 
We used the same 1,000 prompts to generate results from both RoomDiffusion and several open-source models, then had evaluators perform GSB evaluations on the generated results. 
Over 20 evaluators participated in the assessment, covering three dimensions: aesthetic evaluation, text-image alignment, and layout rationality.
\begin{itemize}
    \item[•]\textbf{Aesthetic Evaluation}: Evaluators assessed the images from multiple perspectives, including color coordination, stylistic harmony, and lighting, to select what they considered the best image.
    \item[•]\textbf{Text-Image Alignment}: Evaluators compared the text with the corresponding generated images and selected the one with the highest suitability.
    \item[•]\textbf{Layout Rationality}: Evaluators chose the image with the most reasonable spatial and furniture layout.
\end{itemize}
Each image was evaluated by at least three evaluators, and the final conclusion was based on the majority opinion. 
If a majority opinion could not be reached, the image was excluded from the final statistical results.
    
\section{Results}
\subsection{Machine Indicator Evaluation}
We compared RoomDiffusion with some of the most excellent and widely-used models in the open-source community, such as  EpicRealism,  Realistic Vision, and SDXL. The test set consisted of 1,000 randomly selected interior decoration images, with corresponding text descriptions generated by GPT-4V(ision).
We followed the seven metrics mentioned in Section 3.2, which include aesthetics, object generation success rate, wall-ceiling-floor accuracy, style accuracy, furniture repetition rate, FID, and CLIP score, the comparison results are shown in Table \ref{flare}. Our model achieved the best performance across all machine evaluation metrics; particularly, it excelled in style accuracy and aesthetics scores, while significantly reducing the furniture repetition rate to the lowest level.

\begin{table*}[t!]
\centering
\caption{
The comparison of RoomDiffusion with open-source models. Best results are denoted as \textbf{bold}.}
\footnotesize
{
\begin{tabular}{c|ccccccc}
\hline
 Methods & AS $\uparrow$ & SFR $\uparrow$ & SA $\uparrow$ & HFR $\uparrow$ & FRR $\uparrow$ & FID $\downarrow$ & Clip score $\uparrow$ \\
\hline
Stable diffusion & 73.0 & 43.0 & 35.7 & 48.0 & 27.3 & 47.4 & 17.3\\
EpicRealism & 65.7 & 48.6 & 40.6 & 48.7 & 24.5 & 41.1 & 17.4 \\
Realistic Vision & 67.2 & 48.9 & 41.1 & 50.6 & 29.3 & 36.9 & 17.2 \\
SDXL & 71.7 & 42.2 & 42.9 & 43.0 & 15.2 & 33.5 & 17.4\\
SDXL + refiner & 73.0 & 43.0 & 50.2 & 41.0 & 19.3 & 33.7 &
17.0 \\
\hline
\textbf{RoomDiffusion} & \textbf{78.3} & \textbf{54.3} & \textbf{60.1} & \textbf{54.0} & \textbf{14.5} &  \textbf{33.4} & \textbf{17.8} \\
\hline
\end{tabular}
}
\label{flare}
\end{table*}

\begin{figure}
\centering
\includegraphics[width=4in]{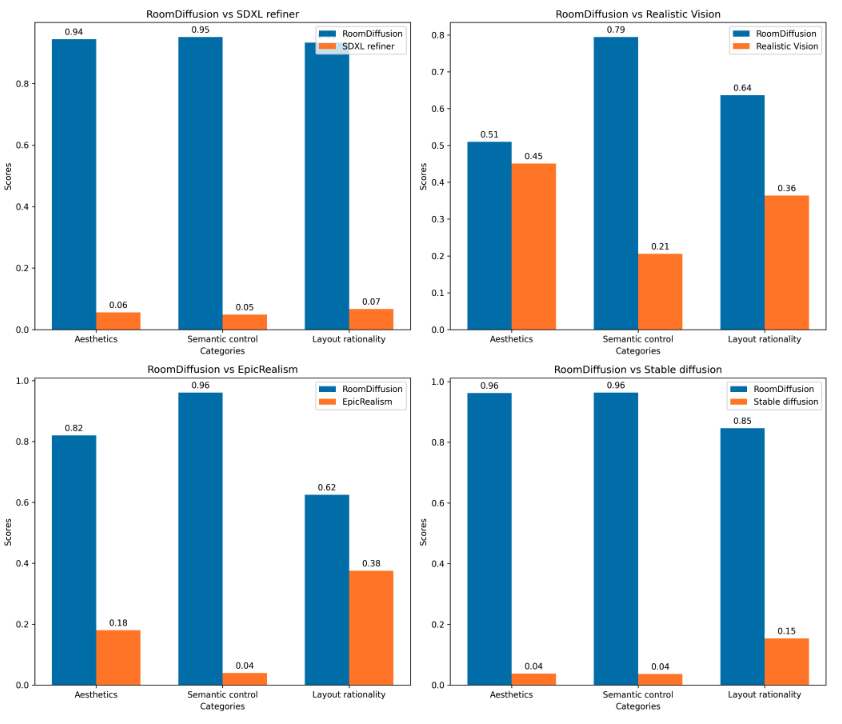}
\caption{Manual evaluation results of RoomDiffusion and open source models}
\label{fig5}
\end{figure}

\subsection{Manual Evaluation}
For manual evaluation, we conducted a comparative assessment based on three dimensions: aesthetics, semantic control, and layout rationality. A total of 20 evaluators participated, comparing the results of RoomDiffusion with those of all the open-source models, and providing their answers from three options: good, same, or bad. After removing the part of "same", the evaluation results are shown in Figure.~\ref{fig5}, where our model demonstrated significant superiority across all dimensions.

\section{Conclusion}
In this report, we introduce RoomDiffusion, an industry model applied to interior decoration design scenarios, which outperforms all existing open-source models. Our report details the construction process of the RoomDiffusion model, the evaluation methods used, and the performance comparison with open-source models. We also hope that our technical report can provide a reference for the open-source community and foster more rapid and valuable development in the field of interior decoration design.
\bibliographystyle{splncs04}
\bibliography{reference}

%




\end{document}